# SADAM: Stochastic Adam, A Stochastic Operator for First-Order Gradient-based Optimizer

**Wei Zhang**
University of California San Francisco
wei.zhang4@ucsf.edu

**Yu Bao**
James Madison University
bao2yx@jmu.edu

## Abstract

In this work, to efficiently help escape the stationary and saddle points, we propose, analyze, and generalize a stochastic strategy performed as an operator for a first-order gradient descent algorithm in order to increase the target accuracy and reduce time consumption. Unlike existing algorithms, the proposed stochastic strategy does not require any batches and sampling techniques, enabling efficient implementation and maintaining the initial first-order optimizer's convergence rate, but provides an incomparable improvement of target accuracy when optimizing the target functions.

In short, the proposed strategy is generalized, applied to Adam, and validated via the decomposition of biomedical signals using Deep Matrix Fitting and another four peer optimizers. The validation results show that the proposed random strategy can be easily generalized for first-order optimizers and efficiently improve the target accuracy.

## 1   Introduction

Optimization problems in machine learning are usually solved by numerical optimizers that either use full gradients, i.e., calculated by the entire dataset, or stochastic gradients, i.e., derived obtained by a single data point or mini-batches of data points. The use of the previous gradient-based optimizer provides guarantees of eventual convergence, and the latter yields advantages in terms of convergence rate and efficient implementation [1-3].

A variety of research works have demonstrated that a hybrid methodology incorporating full gradients, stochastic gradients, noisy stimulus, batch strategy, and sampling techniques can achieve a favorable convergence rate and expected accuracy [4-10]. In short, most first-order gradient-based optimizers generate a sequence as $\{x_t\}_{t=1}^T$ and employs a fundamental formula to update:

$$x_{t+1} \leftarrow x_t + \gamma \cdot f'(x_t) \qquad (1)$$

where $\gamma$ represents a step length and denotes the current derivative. In detail, the full gradients provide variance control for the stochastic gradients, whereas the line of research indicates promising progress towards the goal of designing scalable, autonomous learning algorithms. When solving a nonconvex optimization problem, a full gradient-based optimizer is vulnerable to be trapped in the local optima and/or a saddle point due to their derivatives being equal to zero. Furthermore, stochastic gradient-based optimization is becoming a core practical optimizer in various science and engineering fields since stochastic gradient descent (SGD) provides a mechanism to increase the probability of converging to the global optimum. SGD has been proved as an efficient and effective optimization method since the convergence rate to critical points of a nonconvex problem is $\mathcal{O}(1/T^{1/4})$ [11]. It has also been successfully applied in many machine learning algorithms, such as recent advances in deep learning [12-16]. Objectives may also have other sources of noise than data subsampling, such as dropout regularization [17]. In addition, another peer first-order optimizer employs hybrid strategies, such as stochastic strategy, momentum, and smoothness, to increase the target accuracy. For instance, Stochastic Variance-Reduced Gradient (SVRG) algorithms produce a similar sequence as SGD but adopt the variance-reduced gradient, e.g., a random sampling of previously updated gradient and an averaged gradient of all the current target functions, to further increase the target accuracy and improve the convergence rate to $\mathcal{O}(1/T^{1/3})$ or $\mathcal{O}(1/T^{10/3})$ [18, 19]. In contrast, advanced Adam employs the first, second moment, and adaptive learning rate in order to take advantage of AdaGrad and RMSProp and achieves a convergence rate as $\mathcal{O}(1/T^{1/2})$. Furthermore, Stochastic Recursive Momentum (STORM) proposed a technique of recursive momentum to obtain the convergence rate as $\mathcal{O}(1/T^{2/3})$ [20].

**Contributions**. Although the convergence rate of these first-order optimizers is continuously increasing, there has not been a rapid improvement in the target accuracy level. Therefore, we propose an efficient stochastic strategy as a generalized operator for a first-order gradient-based optimizer.

*Improved Efficiency*. Unlike the stochastic strategies that implement time-consuming sampling techniques [18, 19], our proposed stochastic strategy does not require any sampling techniques or momentum to avoid trapping in the local optima even with the current gradient closing to 0. Instead, the proposed stochastic strategy is implemented via randomizing the target variables.

*Additional Parameters Tuning Free*. As discussed before, the proposed stochastic strategy does not only need the sampling process that is adopted by SVRG, but also does not require the additional momentum that is widely used by Adam and STORM, enabling additional parameters tuning free to our proposed strategy. For instance, there have not been tuning the parameters, such as sampling ratio and learning rate, to control the momentum when employing our proposed stochastic strategy. Briefly, our proposed stochastic method can be implemented in each iteration and a given threshold for initialization. Thus, our proposed stochastic strategy should be easily deployed and applied to any optimizer.

*Significant Generalization*. The proposed stochastic strategy can be easily generalized and applied to most first-order methods to significantly improve the target accuracy level; moreover, we prove that the application of the proposed stochastic strategy is able to maintain the convergence rate of the initial first-order optimizer.

Related Works and Methodological Validation. To validate our proposed stochastic strategy, from an experimental study perspective, we apply our proposed stochastic strategy to one representative method, e.g., Adam [21], to generate Stochastic Adam (SADAM) to compare the target accuracy level, time-consuming, and robustness with the other peer optimizers such as Alternative Direction of Multiplier Method (ADMM) [22], ADAM, and SVRG, based on Deep Matrix Fitting problem. Furthermore, our theoretical analytics demonstrate that our proposed stochastic strategy can guarantee convergence to the global optimum with a larger probability and always maintain the convergence rate of the initial optimization method. Moreover, our theoretical analytics also indicate that the proposed stochastic strategy is superior to other peer optimizers when applied to optimize a Deep Neural Network with extremely complex architecture.

## 2      Assumptions, Definitions, Algorithms, and Analytics

Through this work, we consider the following minimization problem:

$$min_{X \subseteq \mathbb{R}^{M \times N}} F(X) \qquad (2)$$

In Eq. (2), $X$ denotes a matrix as the independent variable of the target function and $F(\cdot)$ denotes a smooth real function. For instance, the matrix functions in this work can be defined as: $\forall k \in \mathbb{N}, F(X) = X^2 \stackrel{\text{def}}{=} XX^T, G(X) = X^3 = (XX^T)^k X, X \in \mathbb{R}^{S \times T}$.

Furthermore, we have three crucial assumptions that will be used in various context in this work as follow:

*Assumption 1* (Finite Dimensionality Space) The real function, mapping, and operators discussed in this work are denoted as, $f: \mathbb{R}^{M \times N} \to \mathbb{R}^{M \times N}$. And the domain of $f$ is a compact set or closed interval $I$.

*Assumption 2* (Norm Equality) [23, 25] Given the finite dimensionality space, all norms are equivalent. In detail, $\|\cdot\|$ represents all norms, such as $\ell_1$ and $\ell_2$ norm.

*Assumption 3* (Banach Space) Given the Assumption 1 and 2, all operators discussed in this work are defined in Banach Space; it means: given an operator $\mathcal{G}: \mathbb{R}^{M \times N} \to \mathbb{R}^{M \times N}$, $\mathcal{G} \in \mathfrak{B}(\mathbb{R}^{M \times N}, \|\cdot\|)$.

*Assumption 4* (Infinite Dimensionality Bounded Space) Given a set $C \in \mathbb{R}^{\infty \times \infty}$, $\|C\| < \infty$.

*Definition 1* (Variance Bounded Real Function) [24] Given a real function $f$ denoted on $[a, b] \subseteq \mathbb{R}$, and $\Delta: a = x_0 < x_1 < x_2 < \cdots < x_n = b$. A sum as $v_\Delta = \sum_{i=1}^{n} |f(x_i) - f(x_{i-1})|$ and $V_a^b(f) = \sup\{v_\Delta : \forall \Delta\}$. The variance bounded real function is denoted as $V_a^b(f) < \infty$.

*Definition 2* (Amplitude of Real Function) [24] Given a real function $f$ denoted on $[a, b]$, and $\forall B(x_0, \delta) \subseteq [a, b], \delta > 0; \omega_f(x_0) = \lim_{\delta \to 0} \sup\{|f(x') - f(x")| : x', x" \in B(x_0, \delta)\}$.

*Definition 3* (Lipschitz Smooth) Given a real differentiable function $f: \mathbb{R}^{M \times N} \to \mathbb{R}^{M \times N}$, $|f(X_1) - f(X_2)| \leq L|X_1 - X_2|$.

*Definition 4* (Lipschitz Gradient Smooth) Given a real differentiable function $f: \mathbb{R}^{M \times N} \to \mathbb{R}^{M \times N}$, $|f'(X_1) - f'(X_2)| \leq L|X_1 - X_2|$.

*Definition 5* (Lipschitz Hessian Smooth) Given a real differentiable function $f: \mathbb{R}^{M \times N} \to \mathbb{R}^{M \times N}$, $|\nabla^2 f(X_1) - \nabla^2 f'(X_2)| \leq L|X_1 - X_2|$.

*Definition 6* (Proposed Stochastic Strategy/Operator in Finite Dimensionality Space) Given a real differentiable function $\mathcal{R}: \mathbb{R}^{M \times N} \to \mathbb{R}^{M \times N}$, $\mathcal{R} \cdot [x_1 \quad x_2 \quad \cdots \quad x_N] = [\hat{x}_1 \quad \hat{x}_2 \quad \cdots \quad \hat{x}_N], \forall x_i \hat{x}_i \in \mathbb{R}^M$ and we have: $\hat{x}_1 = x_i \; \hat{x}_2 = x_j \cdots \hat{x}_n = x_k, i, j, k \in 1, 2, \cdots n$.

*Definition 7* (Proposed Stochastic Strategy/Operator in Infinite Dimensionality Space) Given an operator, $\mathcal{R}: \mathbb{R}^{M \times N} \to \mathbb{R}^{M \times N}$, $\mathcal{R} \cdot [x_1 \quad x_2 \quad \cdots \quad x_k \quad \cdots] = [\hat{x}_1 \quad \hat{x}_2 \quad \cdots \quad \hat{x}_N \quad \cdots] \; \forall x_i, \hat{x}_i \in \mathbb{R}^{M \times 1}$ and we have: $\hat{x}_1 = x_i \; \hat{x}_2 = x_j \cdots \hat{x}_n = x_k \cdots, i, j, k \in 1, 2, \cdots, \infty$.

*Definition 8* (First-Order Operator) Given a first-order operator $\mathcal{G}: \mathbb{R}^{M \times N} \to \mathbb{R}^{M \times N}$, $\|\mathcal{G}X\| \leq \|X\|$ holds.

## 2.1  An Efficient Stochastic Strategy Synergizes Convergence to the Global Optimum

The following section provides the details of the stochastic strategy for the first-order methods to efficiently increase the target accuracy level. In other words, the proposed stochastic strategy expands the domain for gradient to update that allows to increase the probability to converge to the global optimum.

The definition of the proposed stochastic strategy is as follows:

As discussed before, the vital advantage of the proposed stochastic strategy is three folds: 1). The proposed stochastic strategy can increase the probability of converging to the global optimum; 2). This proposed strategy can be generalized to any first-order optimizer to further improve the target accuracy level and even maintain the convergence of the initial convergence rate. 3). The proposed strategy does not require additional parameter tuning and can be easily applied to first-order methods.

Furthermore, we theoretically analyze and explain the advantages of the proposed stochastic strategy. At first, we employ set theory to explain why the stochastic strategy can easily converge to the global optimum that significantly improve the targe accuracy level. All theoretical analytics and proof are based on matrix variable, and based on real analysis and functional analysis.

Given Assumptions 1, and Lemma 1.2, Corollary 1.1, and 1.2 as follows, Theorem 1 concludes that the proposed stochastic operator can increase the probability of converging to the global optimum.

*Lemma 1.2* (Contraction of Operators Combination) Given two contraction mappings $\Phi_1$ and $\Phi_2$, the composite of two contraction mapping as $\Phi_2 \cdot \Phi_1$. The composite mapping $\Phi_2 \cdot \Phi_1$ must be contractive.

*Corollary 1.1* (General Contraction Operator) According to Lemma 1.2, denote the operators $\{\Phi_i\}_{i=1}^K$, $\forall \Phi_i\ i \in \mathbb{N}$, $\Phi_i: \mathbb{R}^{S \times T} \to \mathbb{R}^{S \times T}$; consider any combination of operators: $\Phi_K \cdot \cdots \cdot \Phi_2 \cdot \Phi_1$, if at least a single operator $\Phi_i$ is contraction operator, and the other operators are bounded, such as $\forall i \neq k\ \|\Phi_i\| \leq M$. The combination of operator series $\Phi_K \cdot \cdots \cdot \Phi_2 \cdot \Phi_1$ is a contraction operator, if and only if $\prod_{i=1}^K \|\Phi_i\| < 1$.

*Corollary 1.2* (Iterative Contraction Operator) According to Lemma 1.2, denote the operators $\{\Phi_i\}_{i=1}^K$, $\forall \Phi_i\ i \in \mathbb{N}$, $\Phi_i: \mathbb{R}^{S \times T} \to \mathbb{R}^{S \times T}$; consider any combination of operators: $\Phi_K \cdot \cdots \cdot \Phi_2 \cdot \Phi_1$, if at least a single operator $\Phi_i$ is contraction operator, and other operators are bounded, such as $\forall i \neq k, \|\Phi_i\| \leq M$. The combination of operator series $\Phi_K^n \cdot \cdots \cdot \Phi_2^n \cdot \Phi_1^n$, if and only if $\lim_{n \to \infty} \prod_{i=1}^K \|\Phi_i\|^n = c < 1$.

*Lemma 1.3* (Contraction of First-Order Operators) Given a first-order operator $\mathcal{G}: \mathbb{R}^{M \times N} \to \mathbb{R}^{M \times N}$, $\forall k \in \mathbb{K}$. Consider $\mathcal{G}([x_k, y_k]) = [x_{k+1}, y_{k+1}]$, $[x_{k+1}, y_{k+1}] \subset [x_k, y_k]$ holds.

In Lemma 1.3, consider the iteration as $T$ and $\mathcal{G}^k([x_0, y_0]) = [x_k, y_k]$, $I = [x_0, y_0]$, we can easily conclude: for a series like $\mathcal{G}^1([x_0, y_0]) = [x_1, y_1]$, $\mathcal{G}^1([x_1, y_1]) = [x_2, y_2]$, $\cdots$, $\mathcal{G}^1([x_{k-1}, y_{k-1}]) = [x_k, y_k]$, $\cdots$, $\mathcal{G}^1([x_{T-1}, y_{T-1}]) = [x_T, y_T]$, $[x_0, y_0] \supset [x_1, y_1] \supset [x_2, y_2] \supset \cdots \supset [x_k, y_k] \supset \cdots \supset [x_T, y_T]$ holds. This result indicates that the series of sets/intervals generated via each iteration for first-order methods cannot guarantee the coverage of the initial domain $I$, which might lead to the failure of finding the global optimum: $\{x^*\} \in [x_0, y_0]$ but $\{x^*\} \notin \{[x_1, y_1], [x_2, y_2], \cdots, [x_k, y_k], \cdots, [x_T, y_T]\}$.

The proof of Lemma 1.2, 1.3, and the other two corollaries can be viewed in Appendix A, Supplementary Material. According to Lemma 1.3, any first-order method/operator cannot guarantee the coverage of the initial domain of the target real function, since the intervals generated via the first-order operator have shrunk based on the previous iterations.

This results indicate that the series of set/intervals generated via each iteration for first-order methods cannot guarantee the converge of initial domain $I$, which might lead to the failure of finding the global optimum, given $I = [x_0, y_0]$, $x^* \in [x_0, y_0] \setminus [x_1, y_1]$ but $\{x^*\} \notin \{[x_1, y_1], [x_2, y_2], \cdots, [x_k, y_k], \cdots, [x_T, y_T]\}$. It demonstrates that : if the initialized searching interval does not include the global optimum, the following intervals generated via contraction operators cannt contain the global optimum any more.

Therefore, the natural idea is to utilize a stochastic operator to expand the searching intervals introuduce in the following Theorem 1.4 rather than a contraction operator resulting in a shrinkage of the searching intervals introduced in Lemma 1.3. Originally, due to the contraction of first-order optimizer, we have: $\mathcal{G}^1([x_t, y_t]) \subseteq [x_t, y_t]$; a shrinkage of interval series cannot guarantee their union set to include the global optimum when $I = [x_0, y_0]$, $x^* \in [x_0, y_0] \setminus [x_1, y_1]$; fortunately, by introducing stochastic operator , we may have: $\mathcal{R} \cdot \mathcal{G}^1 \cdot ([x_t, y_t]) \supseteq [x_t, y_t]$ that provides an opportunity to use randomized interval series such as $\{[x_t, y_t]\}_{t=1}^T$ to fully cover original search domain $I = [x_0, y_0]$ to increase the probability to search the global optimum.

Then, we theoretically analyze why the stochastic operator can increase the probability to convergence to the glboal optimum in the following Lemmas and Theorem, using covering theorem.

*Lemma 1.4* (Heine-Borel Covering Theorem) [24] Assume $\Gamma$ is a close and bounded set and $\{g_i\}_{i=1}^K \stackrel{\text{def}}{=} G$ is an open set. Then $\bigcup_{i=1}^K g_i \supseteq \Gamma$, and $\bar{\bar{G}} = \aleph_0$.

*Lemma 1.5* (Vitali Covering Lemma) Assume $\{B_i\}_{i=1}^n$ are closed sets and $\forall B_i \cap B_j = \emptyset, i \neq j, E \subseteq \mathbb{R}$, and $m^*(E) < \infty$, if $m^*(E \setminus \bigcup_{i=1}^n B_i) < \varepsilon, \forall \varepsilon > 0$, holds, $\{B_i\}_{i=1}^n$ defines a Vitali Covering of $E$.

*Theorem 1.4* (Convergence to the Global Optimum using Stochastic Strategy) Given a global optimum point $x^* \in I \subseteq \mathbb{R}^{M \times N}$, a target real function $f: \mathbb{R}^{M \times N} \to \mathbb{R}^{M \times N}$, a first-order operator $\mathcal{G}: \mathbb{R}^{M \times N} \to \mathbb{R}^{M \times N}$, and a stochastic operator $\mathcal{R}: \mathbb{R}^{M \times N} \to \mathbb{R}^{M \times N}$. Assume the maximum iteration is $T$, and the definition domain of a target real function $f$ is $I$, we have:

1). According to Assumption 3 and Lemma 1.5, Heine-Borel Covering Theorem, $\mathcal{R}^t \mathcal{G}^t \cdot I \stackrel{\text{def}}{=} [\hat{x}_t, \hat{y}_t] \subseteq I$ holds, if $\forall\ [\hat{x}_i, \hat{y}_i] \subseteq I\ [\hat{x}_j, \hat{y}_j] \subseteq I\ [\hat{x}_i, \hat{y}_i] \cap [\hat{x}_j, \hat{y}_j] \neq \emptyset$ we have:

$$\{x^*\} \in \bigcup_{t=1}^T [\hat{x}_t, \hat{y}_t] \qquad (3)$$

2). According to Assumption 3 and Lemma 1.5, Vitali Covering Theorem, $\mathcal{R}^t\mathcal{G}^t \cdot I \overset{\text{def}}{=} [\hat{x}_t, \hat{y}_t] \subseteq I$ holds, if $\forall [\hat{x}_i, \hat{y}_i] \subseteq I$ $[\hat{x}_j, \hat{y}_j] \subseteq I$ $[\hat{x}_i, \hat{y}_i] \cap [\hat{x}_j, \hat{y}_j] = \emptyset$ we have:

$$\{x^*\} \in (\bigcup_{t=1}^{T}[\hat{x}_t, \hat{y}_t]) \bigcup E \tag{4}$$

$$\bar{\bar{E}} = \aleph_0$$

In Theorem 1.4, it concludes that the global optimum point $\{x^*\}$ should be included in a series set of $\{[\hat{x}_t, \hat{y}_t]\}_{t=1}^T$ or the union of $\{[\hat{x}_t, \hat{y}_t]\}_{t=1}^T$ and $E$ only contains isolated points. It also demonstrates that the global optimum can be approached given a very large number of iterations.

Moreover, we provide Theorem 1.4 to describe the efficacy of the proposed stochastic strategy. In detail, assume the proposed stochastic strategy as an operator, we can prove that the norm of $\mathcal{R}$ is equal to 1 and does not influence the convergence rate of the initial operator, e.g., a first-order method/optimizer.

*Lemma 1.6* (Norm of Stochastic Strategy in Finite Dimensionality Space) Assume the stochastic strategy is an operator $\mathcal{R}: \mathbb{R}^{M \times N} \to \mathbb{R}^{M \times N}$, $\|\mathcal{R}\| = 1$ holds.

*Lemma 1.7* (Norm of Stochastic Strategy in Infinite Dimensionality Space) Assume the stochastic strategy is an operator $\mathcal{R}: \mathbb{R}^{\infty \times \infty} \to \mathbb{R}^{\infty \times \infty}$, $\|\mathcal{R}\| = 1$ holds, if and only if $\forall X, \hat{X} \in \mathbb{R}^{\infty \times \infty}$, we have: $\|\mathcal{R}X\| < \infty$.

In detail, we denote the stochastic operator in infinite dimensionality space as follows:

$$\mathcal{R} \cdot \begin{bmatrix} x_1 \\ x_2 \\ \vdots \\ x_k \\ \vdots \end{bmatrix} = \begin{bmatrix} \hat{x}_1 \\ \hat{x}_2 \\ \vdots \\ \hat{x}_k \\ \vdots \end{bmatrix} \tag{5}$$

with $\forall X$ and $\hat{X}$ being subsets in an infinite dimensionality space and all their elements being equivalent:

$$\hat{x}_1 = x_i \; \hat{x}_2 = x_j \cdots \hat{x}_k = x_n \cdots \tag{6}$$

To ensure the operator, $X$ and $\hat{X}$ are bounded, we have:

$$x_i \to 0, r_i \to 0$$

$$x_i = \mathcal{O}\left(\frac{1}{n^k}\right), r_i = \mathcal{O}(\frac{1}{n^l}) \tag{7}$$

We can easily conclude:

$$\|\mathcal{R}X\| = \left(\sum_{i=1}^{\infty} \lfloor r_i x_i \rfloor\right) = \frac{1}{n^{(k+l)}} \; k+l > 1 \tag{8}$$

Notably, in an infinite dimensionality space, any norm is not equivalent, that is, $\|\cdot\|$ in Lemma 1.6 is not a generalized norm anymore. Therefore, according to Lemma 1.5 and 1.6, we can conclude:

*Theorem 1.6* (Convergence Rate Maintenances) Given the operator composition $\mathcal{R} \cdot \mathcal{G}$, if $\mathcal{G}$ is a general contraction operator, $\|\mathcal{G}^k X - \mathcal{G}^k Y\| \leq \xi \|X - Y\|$ and $0 < \xi < 1$, we have $\|\mathcal{R}^k \mathcal{G}^k X - \mathcal{R}^k \mathcal{G}^k Y\| \leq \xi \|X - Y\|$.

Theorem 1.6 demonstrates that the composition of stochastic operators and other general contraction operators have the same convergence rate.

In addition, Theorem 2.1 describes the complexity of the fully connected Deep Neural Networks (DNNs).

*Theorem 2.1* (Composition of Non-Smooth Activation Functions) Given a non-smoothed activation function $f_i$ with a single non-smooth point, denoted on $[a,b] \subseteq \mathbb{R}^1$, $f_i \in Lip1([a,b]\backslash\{x_i\})$ $i \in \mathbb{N}$. And the composition of $f_i$ and $f_j$ is represented as $f_{j,i} \overset{\text{def}}{=} f_j(f_i(x))$. $\mathcal{F} \overset{\text{def}}{=} f_{\cdots,k,\cdots j,i} \in Lip1([a,b]\backslash$ $[c, d]$), when $k \to \infty$, and $m([c, d]) \neq 0$. Moreover, when $t \to \infty$, $\sum_{i=1}^{t} \mathcal{F}_t \notin Lip1([a, b] \setminus [c', d'])$ and $m([c', d']) \neq 0$, where $m(\cdot)$ represents the Lebesgue measure.

Theorem 2.1 indicates that a DNN with fully connected architecture and a large number of layers can result in a target function of DNNs that is non-smooth almost everywhere, i.e., $|f(x) - f(y)| > L|x - y|, a.e. x, y \in [a, b]$. Although the optimization of DNNs is challenging, the stochastic strategy guarantees to contain the global optimum within updates, given a large number of iterations, according to Theorem 1.3.

## 2.2 Algorithm

Inspired by the various advantages of the proposed stochastic strategy, naturally, we apply this stochastic method to first-order optimizer to construct an advanced algorithm in a stochastic and efficient fashion to optimize the initial target function.

At first, the proposed stochastic strategy is applied to ADAM optimizer, called SADAM, since ADAM has various achievements in convergence rate and target accuracy level. SADAM is conducted via the fundamental framework of ADAM, and can maintain the initial convergence rate of ADAM which provides an opportunity to significantly increase the target accuracy level. The pseudo-code of SADAM is presented as Algorithm 1 as follows:

---

**Algorithm 1 (SADAM):** Stochastic ADAM

**Input:** $\alpha$ denotes step size; $\beta_1$ and $\beta_2$ are exponential decay; $\theta_0$ is initial parameter vector; $m_0$ is initialized 1st moment vector; $v_0$ is initialized 2nd moment vector; $t$ is initialized as timestep; $\varepsilon$ is initialized as a threshold.

**while** $X_t$ does not converge

$\quad G_t \leftarrow \nabla_X F(X_{t-1})$

$\quad M_t \leftarrow \beta_1 M_{t-1} + (1 - \beta_1) G_t$

$\quad V_t \leftarrow \beta_2 V_{t-1} + (1 - \beta_2) G_t$

$\quad \widehat{M}_t \leftarrow M_t / (1 - \beta_1^t)$

$\quad \widehat{V}_t \leftarrow V_t / (1 - \beta_2^t)$

$\quad X_t \leftarrow X_{t-1} - \alpha \cdot \widehat{M}_t / (\sqrt{\widehat{M}_t} + \epsilon)$

$\quad$ **if** $\|G_{t-1} - G_t\| < \varepsilon$

$\quad\quad \hat{G}_t \leftarrow \mathcal{R}(G_t)$

$\quad\quad G_{t+1} \leftarrow \hat{G}_t$

$\quad$ **end if**

$\quad t \leftarrow t + 1$

**end while**

**Output:** $X_T$ and $F(X_T)$

---

In detail, the difference between ADAM and SADAM is that SADAM employs the proposed stochastic strategy to randomize the updated gradient $G$, which denotes a gradient of the matrix variable. In addition, the stochastic strategy is only applied when the difference between the current and previous gradient reaches the given threshold.

Then, the proposed optimizer SADAM is validated via a multi-layer stacked linear model named Deep Matrix Fitting (Deep MF) to solve the problem of matrix decomposition of biomedical signal matrices [26-29]. This novel Deep MF does not need to tune all hyperparameters [29], which benefits our validations experiments. The formula for the target functions of Deep MF is presented as follows:

$$min_{Z_i \in \mathbb{R}^{m \times n}} \bigcup_{i=1}^{M} \|Z_i\|_1 \qquad (9\text{-}1)$$

$$s.t. \prod_{i=1}^{k} X_i Y_k + Z_k = S \qquad (9\text{-}2)$$

In [29], for Eqs. (9-1) and (9-2), there are three parameters to be optimized. In detail, $X_i$ represents the weight/mixing matrix, $Y_k$ denotes the feature matrix, and $Z_k$ represents the noise/background components. $S$ represents the input matrix. Our proposed SADAM and other three peer optimizers are used to update three variables in Eqs. (9-1) and (9-2). Furthermore, the training loss is defined as $\left\| \prod_{i=1}^{k} X_i Y_k + Z_k - S \right\|_F^2$ [29].

In the following section, all experimental studies including the validations of target accuracy and time consumption based on the decomposition of multiple subjects' signal matrices are presented. Furthermore, some statistical method, such as Intraclass Correlation Coefficient (ICC) is employed to analyze the consistency of SADAM.

## 3  Empirical Validation

As shown in Eqs. (9-1) and (9-2), SADAM and other three peer optimizers are validated on the public biomedical data in Consortium for Neuropsychiatric Phenomics (CNP) (https://openfmri.org/dataset/ds000030/). In this experimental study, all algorithms terminate after two hundred iterations with other parameters fixed to the reported default values in [19, 20, 21]. Moreover, the $\varepsilon$ of difference between the previous and current gradient is fixed as $1 \times 10^{-5}$.

In detail, all optimizers are applied to Deep MF; due to the automatic estimation technique of Deep MF, all subjects' signal matrices are decomposed into two layers. In addition, we provide randomly selected four subjects' training loss curves for the first and second layers after the optimizer stops. Moreover, to decompose all subjects' signal matrices, we also provide the averaged training loss that can further compare the performance of SADAM with other peer algorithms.

In short, SADAM can achieve the highest target accuracy level, considering all subjects and averaged training loss. Moreover, SVRG and SADAM perform better than ADMM and ADAM; when the number of iterations is smaller than one hundred, SVRG can obtain a higher accuracy than SADAM. Nevertheless, when the number of iterations is larger than one hundred, SADAM enables the highest target accuracy level. In other words, this empirical study also proves and validates our theoretical analytics for the stochastic strategy that allows for the increased probability of approximating global optimum.

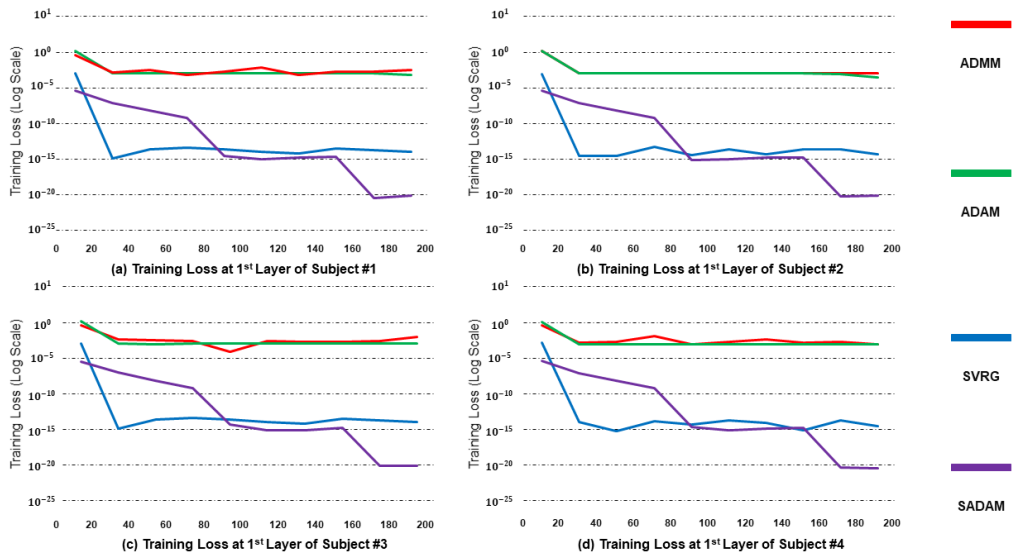

Figure 1. The training loss comparison of proposed SADAM and other three peer optimizers within two hundred iterations of randomly selected four subjects at the first layer.

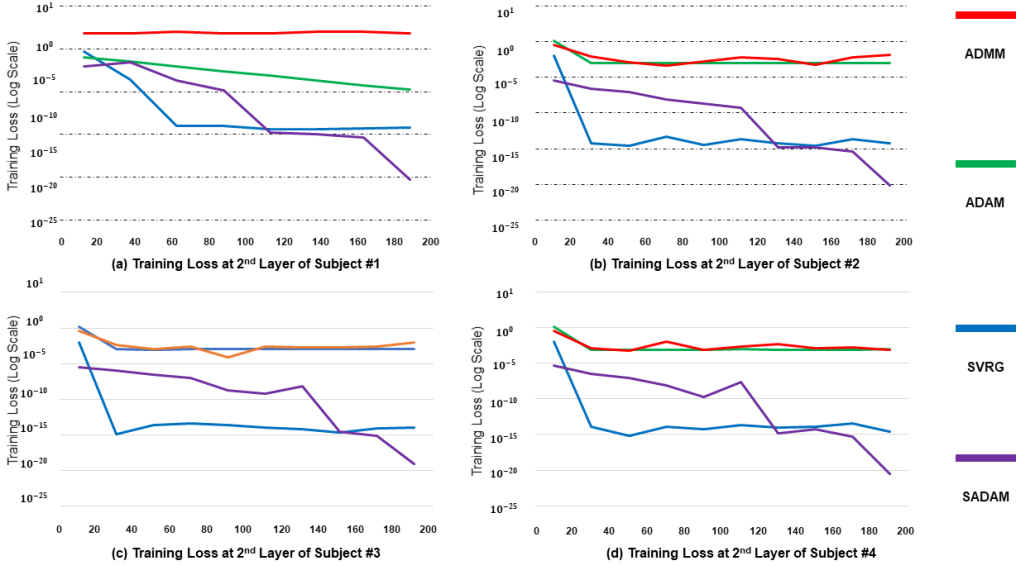

Figure 2. The training loss comparison of proposed SADAM and other three peer optimizers within two hundred iterations of randomly selected four subjects at second layer.

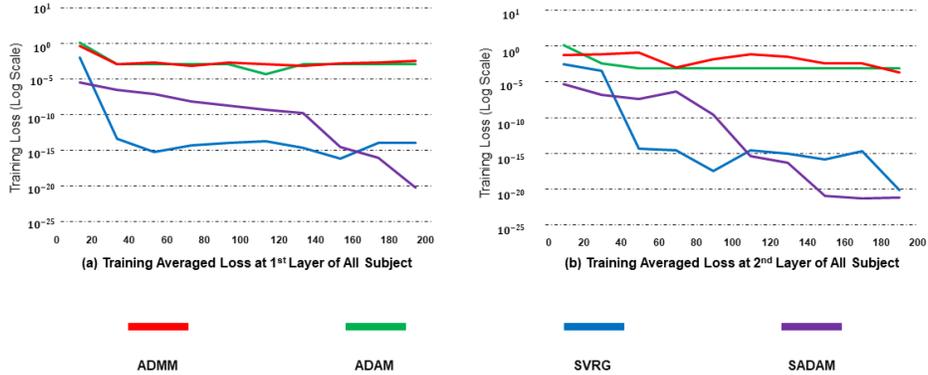

Figure 3. The averaged training loss comparison of proposed SADAM and other three peer optimizers within two hundred iterations of all subjects at first and second layers, respectively.

## 4      Results of Statistical Analytics

Furthermore, we aim to compare the robustness and time consumption of SADAM with other three peer algorithms. In this section, the validation results demonstrate that the time consumption of SADAM is larger than ADMM and ADAM but much smaller than SVRG. Moreover, by calculating the ICCs of training loss using all subjects, it is apparent that SVRG and SADAM are more robust than others. It is also noticeable that SVRG is the most robustness than other peer algorithms since SVRG utilizes the averaged gradient to update. Importantly, without adopting the averaged gradient, SADMA can even perform optimization consistently.

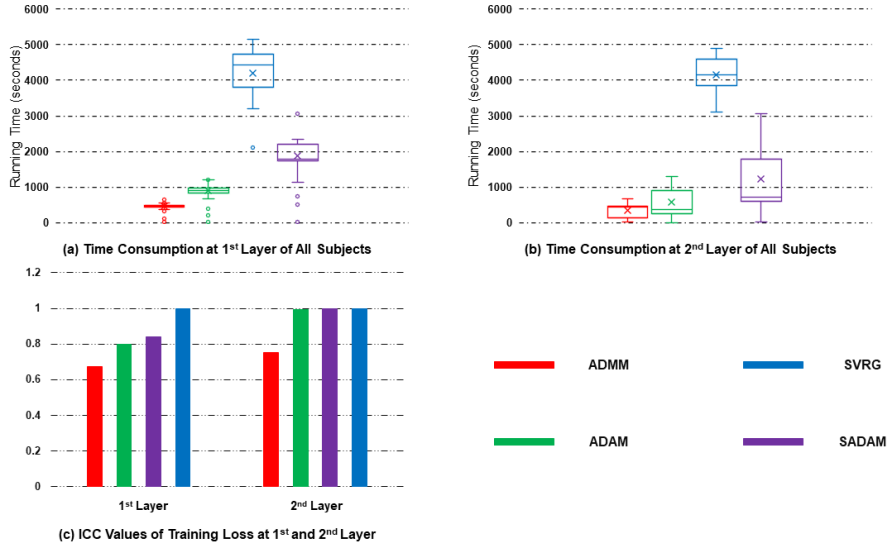

Figure 4. Time-consumption and consistency comparisons of proposed SADAM and other three peer algorithms. In (a) and (b), based on all subjects at the first and second layers, respectively. The box plots represent the time consumptions of four algorithms using all subjects; furthermore, in (c), it provides the ICC values to demonstrate the consistency/robustness of the four algorithms.

In addition, the following Table 1 presents the mean and standard deviation of all time consumptions of four algorithms using all subjects for further validation.

Table 1. Time Consumption Comparison of SADAM and Other Peer Three Optimizers

| Time Consumption at $1^{st}$ Layer | | Time Consumption at $2^{nd}$ Layer | |
|---|---|---|---|
| ADMM | $451.58 \pm 93.86$ | ADMM | $287.42 \pm 78.15$ |
| ADAM | $901.65 \pm 196.96$ | ADAM | $579.98 \pm 53.41$ |
| SVRG | $4538.50 \pm 305.49$ | SVRG | $4441.60 \pm 547.12$ |
| SADAM | $1562.50 \pm 185.31$ | SADAM | $1228.70 \pm 195.09$ |

In Table 1, these results further demonstrate the consistency of SADAM. In detail, only ADMM provides the least time consumption. Due to adding stochastic strategy adopted by SADAM, the time consumption of SADAM is improved and more than ADAM but obviously less than SVRG.

## 5    Conclusion

In this work, we introduce a novel stochastic first-order algorithm that can be applied to stochastic, smooth, and nonconvex problems. We propose a novel algorithm called SADAM by incorporating the stochastic strategy into a prior algorithm ADAM. SADAM further increases the training loss by randomly changing the element gradients of a matrix variable in order to improve the probability of approximating the global optimum. Moreover, the theoretical analytics and empirical validations indicate that SADAM can maintain the convergence rate of the initial optimizer. Finally, our empirical experiments show that SADAM maintains high consistency for both layers of the feature components.

In short, SADAM obtains the optimal convergence rate guarantee, adapting to the level of noise in the problem without knowledge of this parameter. Importantly, our theoretical analytics also demonstrate that the proposed stochastic strategy can be easily generalized as a technique for other first-order optimizers. Since it is challenging to improve the convergence rate and the approximation accuracy to global optimum when training DNNs with deep layers [30-31], a future investigation could focus on incorporating the proposed stochastic strategy into other optimizers to validate the improved training loss and the maintenance of the convergence rates of the initial optimizers.

# Supplemental Material

# Appendix A

***Assumption 1.1*** For any operator discussed in this study, we have: $\forall \mathcal{C} \in \mathfrak{C}, \mathcal{C}: \mathbb{R}^{S \times T} \to \mathbb{R}^{S \times T}, S < \infty, T < \infty$. This assumption demonstrates that all operators are mapping from the finite dimensional space to another finite dimensional space, which is also reasonable in the real world.

***Lemma 1.1* (Norm Equality)** Given any arbitrary norm $\|\cdot\|$ and/or their finite linear combination $\sum_{i=1}^{n} k_i \|\cdot\|$ denoted based on any finite set, this norm or their finite linear combination is equivalent to $\ell_2$ norm (e.g., $\|\cdot\|_2$).

***Lemma 1.2* (Contraction of Operators Combination)** Given two contraction mappings $\Phi_1$ and $\Phi_2$, we have the composite of two contraction mapping as $\Phi_2 \cdot \Phi_1$. The composite mapping $\Phi_2 \cdot \Phi_1$ must be contractive.

*Proof*: According to the definition of contraction linear operator, we have:

$$\exists \zeta \in (0,1)$$

$$\rho \overset{\text{def}}{=} \|\Phi x - \Phi y\| \tag{A.1}$$

$$\rho(\Phi x, \Phi y) \leq \zeta \rho(x, y)$$

Obviously, and we have:

$$\rho(\Phi_1 u, \Phi_1 v) \leq \zeta \rho(u, v) \;\; \forall \zeta \in (0,1) \tag{A.2}$$

$$\rho(\Phi_2 x, \Phi_2 y) \leq \eta \rho(x, y) \;\; \forall \eta \in (0,1)$$

If we set:

$$x = \Phi_1 u, y = \Phi_1 v \tag{A.3}$$

the inequality below holds:

$$\rho(\Phi_2 x, \Phi_2 y) \leq \eta \rho(\Phi_1 u, \Phi_1 v) \leq \zeta \eta \rho(u, v) \tag{A.4}$$

Since the definition as

$$\forall \zeta, \eta \in (0,1), \rho(\Phi_2 \Phi_1 u, \Phi_2 \Phi_1 y) \leq \zeta \eta \rho(u, v) \tag{A.5}$$

***Corollary 1.1* (General Contraction Operator)** According to Lemma 1.2, if denote the operators $\{\Phi_i\}_{i=1}^{K}$, $\forall \Phi_i \; i \in \mathbb{N}, \Phi_i: \mathbb{R}^{S \times T} \to \mathbb{R}^{S \times T}$; considering any combination of operators: $\Phi_K \cdot \cdots \cdot \Phi_2 \cdot \Phi_1$, if at least a single operator $\Phi_i$ is contraction operator, and other operators are bounded, such as $\forall i \neq k \; \|\Phi_i\| \leq M$. If and only if $\prod_{i=1}^{K} \|\Phi_i\| < 1$, the combination of operator series $\Phi_K \cdot \cdots \cdot \Phi_2 \cdot \Phi_1$ is a contraction operator.

*Proof*: Obviously, according to Lemma 1.2, use a series as $\{\zeta_i\}_{i=1}^{K}$ to replace $\zeta, \eta \in (0,1)$,

Obviously, we have:

$$\zeta_i \in (0,1) \; i \in \mathbb{N} \tag{A.6}$$

$$\rho(\Phi_K \cdot \cdots \cdot \Phi_2 \Phi_1 u, \Phi_K \cdot \cdots \cdot \Phi_2 \Phi_1 y) \leq \zeta_K \cdot \cdots \zeta_2 \cdot \zeta_1 \cdot \rho(u,v)$$

Since $\zeta_K \cdot \cdots \zeta_2 \cdot \zeta_1 < 1$, we have proved this corollary.

*Corollary 1.2* (**Iterative Contraction Operator**) According to Lemma 1.2, if denote the operators $\{\Phi_i\}_{i=1}^{K}$, $\forall \Phi_i \; i \in \mathbb{N}$, $\Phi_i: \mathbb{R}^{S \times T} \to \mathbb{R}^{S \times T}$; considering any combination of operators: $\Phi_K \cdot \cdots \cdot \Phi_2 \cdot \Phi_1$, if at least a single operator $\Phi_i$ is contraction operator, and other operators are bounded, such as $\forall i \neq k, \|\Phi_i\| \leq M$. If and only if $\lim_{n \to \infty} \prod_{i=1}^{K} \|\Phi_i\|^n = c < 1$, the combination of operator series $\Phi_K^n \cdot \cdots \cdot \Phi_2^n \cdot \Phi_1^n$.

*Proof:* Obviously, according to Lemma 1.2 and Corollary 1.1 and 1.2, use a series as $\{\zeta_i\}_{i=1}^{K}$ to replace $\zeta, \eta \in (0,1)$,

And we have:

$$\forall \zeta_i \in (0,1) \; i \in \mathbb{N} \tag{A.7}$$

$$\rho(\Phi_K^n \cdot \cdots \cdot \Phi_2^n \cdot \Phi_1^n u, \Phi_K^n \cdot \cdots \cdot \Phi_2^n \cdot \Phi_1^n y) < \zeta_i^n \cdot \cdots \cdot \zeta_2^n \cdot \zeta_1^n \cdot \rho(u,v)$$

Since $0 < \zeta_i^n \cdot \cdots \cdot \zeta_2^n \cdot \zeta_1^n < 1$, we have proved this corollary.

*Theorem 1.1* (**Contraction of ADMM Operator**) ADMM could be considered as contraction operator. It can be treated as a general iterative contraction operator in finite dimensionality space. We have $ADMM \stackrel{\text{def}}{=} \mathcal{A}$. If denote the $\|\mathcal{A}^{k+1}\| = \alpha \|\mathcal{A}^k\|$, and $\beta$ should be step length, i.e., penalty parameter, if $n \to \infty \; 0 < (\alpha \beta)^n \|BN\| < 1$, $\mathcal{A}$ can be considered as a contraction operator. And $\|BN\|$ denotes the norm of different residual error, considering two distinctive input matrices.

*Proof:* $X$ and $Y$, represent the two input matrices.

Consider the iterative format of ADMM as

$$\mathcal{A}_{k+1} \leftarrow \mathcal{A}_k - \min(f_\mathcal{A}) \tag{A.8}$$

And it also can imply:

$$\|\mathcal{A}_{k+1}\| = \alpha \|\mathcal{A}_k\|, \tag{A.9}$$

$$0 < \alpha < 1$$

According to the definition of contraction operator, we have:

$$\|\mathcal{A}X - \mathcal{A}Y\| \leq \alpha \left\| \left( \beta(e_k^t + \prod_{i=1}^{k-1} X_i Y_k + \sum_{i=1}^{k} Z_k^{t+1} - S) - \alpha\beta(\hat{e}_k^t + \prod_{i=1}^{k-1} \hat{X}_i \hat{Y}_k \right. \right.$$
$$\left. \left. + \sum_{i=1}^{k} \hat{Z}_k^{t+1} - \hat{S}) \right) \right\| \tag{A.10}$$

And we also have:

$$\|\mathcal{A}X - \mathcal{A}Y\| \leq \alpha\beta \left\| e_k^t - \hat{e}_k^t + \prod_{i=1}^{k-1} X_i Y_k - \prod_{i=1}^{k-1} \hat{X}_i \hat{Y}_k + \sum_{i=1}^{k} Z_k^{t+1} - \sum_{i=1}^{k} \hat{Z}_k^{t+1} + \hat{S} - S \right\| \quad (A.11)$$

Since $e_k^t, \hat{e}_k^t, \prod_{i=1}^{k-1} X_i Y_k, \prod_{i=1}^{k-1} \hat{X}_i \hat{Y}_k, \sum_{i=1}^{k} Z_k^{t+1}, \sum_{i=1}^{k} \hat{Z}_k^{t+1}, \hat{S}, S \in \mathbb{R}^{m \times n}$, they are obviously bounded; and using Corollary 1.1 and 1.2, we have:

$$\|\mathcal{A}X - \mathcal{A}Y\| \leq \alpha\beta \|BN\| \quad (A.12)$$

Obviously, it demonstrates:

$$\|\mathcal{A}^n A - \mathcal{A}^n B\| \leq (\alpha\beta)^n \|BN\| < 1 \quad (A.13)$$

If and only if $0 < (\alpha\beta)^n < 1$, or $0 < \alpha\beta < 1$, $\mathfrak{A}^n$ is equivalent to a contraction operator. According to Lemma 1.2 and Corollary 1.1, 1.2, it also indicates: when $n$ is large enough, $n > N$, we have:

$$\lim_{n \to \infty} \|\mathcal{A}^n A - \mathcal{A}^n B\| \leq \lim_{n \to \infty} (\alpha\beta)^n \|BN\| \quad (A.14)$$

Obviously, if and only if $\lim_{n \to \infty} (\alpha\beta)^n \|BN\| < 1$, the iterative ADMM operator can be equivalent to a contraction operator.

***Theorem 1.2*** **(Contraction of GD Operator)** Gradient Descent (GD) is a bounded contraction operator, if and only if the derivative of target function is bounded: $|f''(\varsigma)| < \frac{1}{\sigma} < \infty$, $\sigma$ is the step length.

*Proof*: The standard iteration format is:

$$x_{k+1} = x_k - \sigma f'(x_k) \quad (A.15)$$

Using the definition of operator, we have:

$$\tau(x_k) = x_k - \sigma f'(x_k) \; \forall \sigma \in (0,1) \quad (A.16)$$

And we have:

$$\|\tau X - \tau Y\| = \|(X - Y) - \sigma(f'(X) - f'(Y))\| \quad (A.17)$$

Using Mean value theorem, we have:

$$\|\tau X - \tau Y\| = |1 - \sigma f''(\varsigma)| \|X - Y\| \quad (A.18)$$

According to the definition of contraction operator [30], if and only if:

$$|1 - \sigma f''(\varsigma)| < 1, |1 - \sigma f''(\varsigma)| \in \mathbb{K} \quad (A.19)$$

It also implies, when the following inequality holds:

$$|f''(\varsigma)| < \frac{1}{\sigma} < \infty \quad (A.20)$$

GD is considered as a contraction mapping/operator. Without generality, we can set $\sigma < \frac{1}{|f''(x)|+1}$.

And obviously, using multiplicative inequality, we have:

$$\|\tau X - \tau Y\| \leq \|\tau\|\|X - Y\| \tag{A.21}$$

Since $X$ and $Y$ both denote in finite $\ell^2$ space, we have:

$$\|\tau\|\|X - Y\| \leq \infty \tag{A.22}$$

Using Uniformly bounded theorem, we have:

$$\|\tau\| \leq M, M \in \mathbb{K} \tag{A.23}$$

GD is a bounded mapping/operator.

According to Lemma 1.2, and Corollary 1.1-1.2, obviously, for $n$ iterations for an operator, and if we set the accuracy level as $\varepsilon$, we have:

$$\|\tau^n X - \tau^{n+1} Y\| = \sigma^n \|X - \tau Y\| < \varepsilon \tag{A.24}$$

Since $X$ and $Y$ is both denoted in finite $\ell^2$ space, we have:

$$\sigma^n \|X - \tau Y\| \leq \sigma^n (\|X\| + \|\tau Y\|) \tag{A.25}$$

Obviously, $\|X - Y\|_{\ell^2}$ is bounded, and we have:

$$\sigma^n (\|X\| + \|\tau Y\|) \leq \sigma^n (\|X\| + \|\tau\|\|Y\|) \leq \sigma^n (\|X\| + \|Y\|) \leq \sigma^n \cdot 2\|X\|$$

$$0 < \sigma^n \cdot 2\|X\| < \varepsilon \tag{A.26}$$

$$n > \log \frac{\varepsilon}{2\|X\|} / \log \sigma > 0$$

We provide the infimum of iteration as $\log \frac{\varepsilon}{2\|X\|} / \log \sigma$ to approach the accuracy level $\varepsilon$.

***Theorem 1.3*** **(Contraction of Adam Operator)** ADAM is a bounded contraction operator.

*Proof*: According to definition of contraction operator, we have:

$$\|\mathcal{A}^t X - \mathcal{A}^{t-1} X\| \leq \left\| \alpha \frac{\widehat{m}_t}{\sqrt{\widehat{v}_t} + \epsilon} \right\| \|X\| \tag{A.27}$$

Then, we can conclude the following inequality:

$$\alpha \left\| \frac{\widehat{m}_t}{\sqrt{\widehat{v}_t} + \epsilon} \right\| \leq \alpha \left\| \frac{\widehat{m}_t}{\sqrt{\widehat{v}_t}} \right\| \tag{A.28}$$

The inequality (A.28) can be rewritten as follow:

$$\left\| \frac{\widehat{m}_t}{\sqrt{\widehat{v}_t}} \right\| = \left\| \frac{\sqrt{1 - \beta_2^t} \cdot m_t}{\sqrt{v_t} \cdot (1 - \beta_1^t)} \right\| \leq \left\| \frac{m_t}{\sqrt{v_t}} \right\| \stackrel{\text{def}}{=} \mathcal{G}^t \tag{A.29}$$

According to the definition of contraction operator, we have:

$$\|\mathcal{G}^t X - \mathcal{G}^{t+k} X\| = \left\|\frac{m_t}{\sqrt{v_t}} - \frac{m_{t+k}}{\sqrt{v_{t+k}}}\right\| \leq \left\|\frac{1}{\sqrt{v_t \cdot v_{t+k}}}\right\| \cdot \left\|\sqrt{v_{t+k}} \cdot m_t - \sqrt{v_t} \cdot m_{t+k}\right\| \quad (A.30)$$

Then, we can derive the following formula:

$$\left\|\frac{1}{\sqrt{v_t \cdot v_{t+k}}}\right\| \cdot \left\|\sqrt{v_{t+k}} \cdot m_t - \sqrt{v_{t+k}} \cdot m_{t+k} + \sqrt{v_{t+k}} \cdot m_{t+k} - \sqrt{v_t} \cdot m_{t+k}\right\| \quad (A.31)$$

Eq (A.31) can be rewritten as:

$$\left\|\frac{1}{\sqrt{v_t \cdot v_{t+k}}}\right\| \cdot \left(\left\|\sqrt{v_{t+k}} \cdot m_t - \sqrt{v_{t+k}} \cdot m_{t+k}\right\| + \left\|\sqrt{v_{t+k}} \cdot m_t - \sqrt{v_t} \cdot m_{t+k}\right\|\right) \quad (A.32)$$

We can easily conclude:

$$\left\|\frac{1}{\sqrt{v_t}}\right\| \cdot \|m_t - m_{t+k}\| + \left\|\frac{1}{\sqrt{v_t \cdot v_{t+k}}}\right\| \cdot \left\|\sqrt{v_{t+k}} \cdot m_t - \sqrt{v_t} \cdot m_{t+k}\right\| \quad (A.33)$$

Then we have:

$$\left\|\frac{1}{\sqrt{v_t \cdot v_{t+k}}}\right\| \cdot \left\|\sqrt{v_{t+k}} \cdot m_t - \sqrt{v_t} \cdot m_{t+k}\right\| \leq \left\|\frac{1}{\sqrt{\min(v_t, v_{t+k})}}\right\| \cdot \|m_t - m_{t+k}\| \quad (A.34)$$

And we can also conclude:

$$\left\|\frac{1}{\sqrt{v_t \cdot v_{t+k}}}\right\| \cdot \left\|\sqrt{v_{t+k}} \cdot m_t - \sqrt{v_t} \cdot m_{t+k}\right\| \leq \left\|\frac{1}{\sqrt{\min(v_t, v_{t+k})}}\right\| \cdot \|m_t - m_{t+k}\| \quad (A.35)$$

Considering another momentum used in ADAM:

$$v_t \stackrel{\text{def}}{=} \beta_2 v_{t-1} + (1 - \beta_2)[f'(x)]^2 \quad (A.36)$$

Similarly, we have:

$$v_t = \beta_2^{t-1} v_0 + \beta_2^{t-1}(1 - \beta_2)[f'(x)]^2 + (1 - \beta_2)[f'(x)]^2 \quad (A.37)$$

We have following equation hold:

$$v_t = \beta_2^{t-1} v_0 + (1 - \beta_2)[f'(x)]^2 \cdot \sum_{i=1}^{t-1} \beta_2^i + (1 - \beta_2)[f'(x)]^2 \quad (A.38)$$

The second momentum can also be rewritten as:

$$v_t = (1 - \beta_2)[f'(x)]^2 \cdot C + (1 - \beta_2)[f'(x)]^2 \quad (A.39)$$

The norm of second momentum is bounded as:

$$\|v_t\| \leq [(1 - \beta_2) \cdot C + (1 - \beta_2)] \cdot \|[f'(x)]^2\| < \infty \quad (A.40)$$

Consider the difference of $m_t$ and $m_{t+k}$:

$$C \cdot \|m_t - m_{t+k}\| \quad (A.41)$$

Obviously, we have:

$$\|m_t X - m_{t+k} X\| = \left\|(1-\beta_2) \sum_{i=t+1}^{t+k} \beta_2^i \cdot f'(x)\right\| \|X\| \qquad (A.42)$$

$$\leq (1-\beta_2) \cdot \sum_{i=t+1}^{t+k} \beta_2^i \cdot \|f'(x)\| \cdot \|X\|$$

The Eq. (A.42) indicates:

$$(1-\beta_2) \cdot \sum_{i=t+1}^{t+k} \beta_2^i \cdot \|f'(x)\| \cdot \|X\| = \frac{1}{\hat{C}_k} \cdot \|X\| \qquad (A.43)$$

Then we can conclude:

$$0 < \prod_{i=1}^{k} \frac{1}{\hat{C}_k} < 1 \qquad (A.44)$$

***Lemma 1.3*** (Contraction of First-Order Operators) Given a first-order operator denoted as $\mathcal{G}: \mathbb{R}^{M \times N} \to \mathbb{R}^{M \times N}$, $\forall k \in \mathbb{K}$, considering $\mathcal{G}([x_k, y_k]) = [x_{k+1}, y_{k+1}]$, we have $[x_{k+1}, y_{k+1}] \subset [x_k, y_k]$ hold.

*Proof:* Proof by contradiction. Assume $[x_{k+1}, y_{k+1}] \supseteq [x_k, y_k]$, it indicates:

$$\mathcal{G}([x_k, y_k]) \supseteq [x_k, y_k] \qquad (A.45)$$

Consider $k \to \infty$, obviously, $\mathcal{G}([x_k, y_k]) \to \infty$ holds.

This conclusion contradicts $\mathcal{G}$ is a contraction operator, according to Theorems 1.1 to 1.3.

***Lemma 1.4*** (**Vitali Covering Lemma**) Given $\{B_i\}_{i=1}^n$ are closed sets and $\forall B_i \cap B_j = \emptyset, i \neq j, E \subseteq \mathbb{R}$, and $m^*(E) < \infty$, if $m^*(E \setminus \cup_{i=1}^n B_i) < \varepsilon, \forall \varepsilon > 0$, holds, $\{B_i\}_{i=1}^n$ defines a Vitali Covering of $E$.

***Lemma 1.5*** (**Heine-Borel Covering Theorem**) Given $\Gamma$ is a close and bounded set. Then, an open set sequence as $\{g_i\}_{i=1}^K \stackrel{\text{def}}{=} G$, $\cup_{i=1}^K g_i \supseteq \Gamma$, and $\bar{\bar{G}} = \aleph_0$.

***Theorem 1.4*** (Convergence to the Global Optimum using Stochastic Strategy) Given a global optimum point denoted as $x^* \in I \subseteq \mathbb{R}^{M \times N}$, a target real function $f: \mathbb{R}^{M \times N} \to \mathbb{R}^{M \times N}$, a first-order operator $\mathcal{G}: \mathbb{R}^{M \times N} \to \mathbb{R}^{M \times N}$, and a stochastic operator $\mathcal{R}: \mathbb{R}^{M \times N} \to \mathbb{R}^{M \times N}$, and considering the maximum iteration as $T$, $I$ denotes the definition domain of a target real function $f$, we have:

1). According to Assumption 3 and Lemma 1.5, Heine-Borel Covering Theorem, $\mathcal{R}^t \mathcal{G}^t \cdot I \stackrel{\text{def}}{=} [\hat{x}_t, \hat{y}_t] \subseteq I$ holds, if $\forall \; [\hat{x}_i, \hat{y}_i] \subseteq I \; [\hat{x}_j, \hat{y}_j] \subseteq I \; [\hat{x}_i, \hat{y}_i] \cap [\hat{x}_j, \hat{y}_j] \neq \emptyset$ we have:

$$\{x^*\} \in \bigcup_{t=1}^{T} [\hat{x}_t, \hat{y}_t] \qquad (A.46)$$

2). According to Assumption 3 and Lemma 1.5, Vitali Covering Theorem, $\mathcal{R}^t \mathcal{G}^t \cdot I \overset{\text{def}}{=} [\hat{x}_t, \hat{y}_t] \subseteq I$ holds, if $\forall [\hat{x}_i, \hat{y}_i] \subseteq I \ [\hat{x}_j, \hat{y}_j] \subseteq I \ [\hat{x}_i, \hat{y}_i] \cap [\hat{x}_j, \hat{y}_j] = \emptyset$, and we have:

$$\{x^*\} \in (\bigcup_{t=1}^{T} [\hat{x}_t, \hat{y}_t]) \bigcup E \tag{A.47}$$

$$\bar{\bar{E}} = \aleph_0$$

*Proof*: According to Assumption 3, we have:

$$\mathcal{R} \cdot \mathcal{G} \cdot I \subseteq I \tag{A.48}$$

Similarly, we also have:

$$\mathcal{G}^t \cdot I \overset{\text{def}}{=} [x_t, y_y] \subseteq I \tag{A.49}$$

And, we can easily obtain:

$$\mathcal{R}^t \mathcal{G}^t \cdot I \overset{\text{def}}{=} [\hat{x}_t, \hat{y}_t] \subseteq I \tag{A.50}$$

For $t \to \infty$, we have:

$$\mathcal{R}^t \mathcal{G}^t \cdot I \overset{\text{def}}{=} [\hat{x}_t, \hat{y}_t] \subseteq I \tag{A.51}$$

$$t \to \infty$$

At first, if we consider the overlap through all generated intervals via $\mathcal{R}^t \mathcal{Q}^t \cdot I$:

$$\forall [\hat{x}_i, \hat{y}_i] \subseteq I \ [\hat{x}_j, \hat{y}_j] \subseteq I \ [\hat{x}_i, \hat{y}_i] \cap [\hat{x}_j, \hat{y}_j] \neq \emptyset \tag{A.52}$$

According Lemma 1.4, if we denote the global optimum point as $\{x_{gbest}\}$, we have:

$$\{x_{gbest}\} \subseteq \bigcup_{t=1}^{\infty} [\hat{x}_t, \hat{y}_t] \tag{A.53}$$

We can also have:

$$f(x_{gbest}) \subseteq f(\bigcup_{t=1}^{\infty} [\hat{x}_t, \hat{y}_t]) \tag{A.54}$$

Furthermore, if we consider no overlap through all generated intervals via $\mathcal{R}^t \mathcal{Q}^t \cdot I$:

$$\forall [\hat{x}_i, \hat{y}_i] \subseteq I \ [\hat{x}_j, \hat{y}_j] \subseteq I \ [\hat{x}_i, \hat{y}_i] \cap [\hat{x}_j, \hat{y}_j] = \emptyset \tag{A.55}$$

According to Lemma 1.5, we have:

$$m^*(I \setminus \bigcup_{t=1}^{\infty} [\hat{x}_t, \hat{y}_t]) < \varepsilon$$

$$\{x_{gbest}\} \subseteq \bigcup_{t=1}^{\infty} [\hat{x}_t, \hat{y}_t] \tag{A.56}$$

$$\{x_{gbest}\} \subseteq I \backslash \bigcup_{t=1}^{\infty}[\hat{x}_t, \hat{y}_t]$$

It also indicates:

$$\{x_{gbest}\} \subseteq (\bigcup_{t=1}^{\infty}[\hat{x}_t, \hat{y}_t]) \bigcup E \qquad (A.57)$$

$$\bar{\bar{E}} = \aleph_0$$

***Lemma 1.6*** (Norm of Stochastic Strategy in Finite Dimensionality Space) Given Stochastic strategy as an operator denoted as $\mathcal{R}: \mathbb{R}^{M \times N} \to \mathbb{R}^{M \times N}$, we have $\|\mathcal{R}\| = 1$ hold.

*Proof*: When operator $\mathcal{R}$ applied on finite dimensionality space, we have:

$$\mathcal{R} \cdot \begin{bmatrix} x_1 \\ x_2 \\ \vdots \\ x_n \end{bmatrix} = \begin{bmatrix} \hat{x}_1 \\ \hat{x}_2 \\ \vdots \\ \hat{x}_n \end{bmatrix} \qquad (A.58)$$

Obviously, we have:

$$\hat{x}_1 = x_i \; \hat{x}_2 = x_j \cdots \hat{x}_n = x_k \qquad (A.59)$$

Eq (A.49) indicates:

$$\|X\| = \|\hat{X}\| \qquad (A.60)$$

According to the definition of operator norm:

$$\|\mathcal{R}\| = \sup \frac{\|\mathcal{R}X\|}{\|X\|} = \sup \frac{\|\hat{X}\|}{\|X\|} = 1 \qquad (A.61)$$

***Lemma 1.7*** (Norm of Stochastic Strategy in Infinite Dimensionality Space) Given Stochastic strategy as an operator denoted as $\mathcal{R}: \mathbb{R}^{\infty \times \infty} \to \mathbb{R}^{\infty \times \infty}$, $\|\mathcal{R}\| = 1$ holds, if and only if $\forall X, \hat{X} \in \mathbb{R}^{\infty \times \infty}$, we have: $\|\mathcal{R}X\| < \infty$.

*Proof*: Similarly, consider an operator $\mathcal{R}$ applied on infinite dimensionality space:

$$\mathcal{R} \cdot \begin{bmatrix} x_1 \\ x_2 \\ \vdots \\ x_k \\ \vdots \end{bmatrix} = \begin{bmatrix} \hat{x}_1 \\ \hat{x}_2 \\ \vdots \\ \hat{x}_k \\ \vdots \end{bmatrix} \qquad (A.62)$$

Obviously, we have:

$$\hat{x}_1 = x_i \; \hat{x}_2 = x_j \cdots \hat{x}_k = x_n \cdots \qquad (A.63)$$

We consider infinite dimensionality space but bounded such as:

$$\|X\| = \|\hat{X}\| < \infty \tag{A.64}$$

Easily, the following inequality holds:

$$\sum_{i=1}^{\infty} \lfloor x_i \rfloor^p < \infty \tag{A.65}$$

Therefore, we have the following conclusions:

$$x_i \to 0$$

$$x_i = \mathcal{O}(\frac{1}{n^k}) \tag{A.66}$$

$$\lfloor x_i \rfloor^p = \frac{1}{n^{kp}} \quad kp > 1$$

***Theorem 1.4*** (Convergence Rate Maintenances) Given the operator composition as $\mathcal{R} \cdot \mathcal{G}$, if $\mathcal{G}$ is a general contraction operator, such as $\|\mathcal{G}^k X - \mathcal{G}^k Y\| \leq \xi \|X - Y\|$ and $0 < \xi < 1$, we have $\|\mathcal{R}^k \mathcal{G}^k X - \mathcal{R}^k \mathcal{G}^k Y\| \leq \xi \|X - Y\|$.

*Proof*: According to Theorems 1.1 to 1.3, all optimizers involved in this paper can be treated as a general contraction operator, such as:

$$\|\mathcal{G}^k X - \mathcal{G}^k Y\| \leq \xi \|X - Y\| \tag{A.67}$$

Given the application of stochastic operator, for each iteration, we have:

$$\|\mathcal{R}^k \mathcal{G}^k X - \mathcal{R}^k \mathcal{G}^k Y\| \tag{A.68}$$

According to norm inequality, we have:

$$\|\mathcal{R}^k \mathcal{G}^k X - \mathcal{R}^k \mathcal{G}^k Y\| \leq \|\mathcal{R}^k\| \cdot \|\mathcal{G}^k\| \cdot \|X - Y\| \tag{A.69}$$

And, according to Eq. (A.69) and Lemma 1.6, we have:

$$\|\mathcal{R}^k \mathcal{G}^k X - \mathcal{R}^k \mathcal{G}^k Y\| \leq \xi \cdot \|X - Y\| \tag{A.70}$$

Therefore, Eq. (A.70) demonstrates that stochastic operator does not influence the convergence rate of initial algorithm.

# Appendix B

***Lemma 2.1*** **(Composition of Function)** Given $g \in Lip1([a,b])$, and $g$ is not a constant real function, if $f \notin Lip1([a,b])$, $g(f(x)) \notin Lip1([a,b])$ holds.

*Proof*: Proof by contradiction, if assume $g(f(x)) \in Lip1([a,b])$, $\forall x_1, x_2 \in [a,b]$, $f(x_1), f(x_2) \in [a,b]$,

$$|g(f(x_1)) - g(f(x_2))| < L_g|f(x_1) - f(x_2)| < N < \infty \tag{B.1}$$

$$L_g \neq 0$$

However, since $f \notin Lip1([a,b])$, we have: $|f(x_1) - f(x_2)| > M$ that is contradiction with $|f(x_1) - f(x_2)| < \frac{N}{L_g}$. Thus, $(f(x)) \notin Lip1([a,b])$ holds.

***Lemma 2.2*** **(Smooth & Variance Bounded Real Function)** If and only if a real function $f \in Lip1([a,b])$, $V_a^x(f) < \infty$ holds.

*Proof*: If $f \in Lip1([a,b])$, it indicates: $\forall x_1, x_2 \in [a,b]$, we have: $|f(x_1) - f(x_2)| \leq L|x_1 - x_2|$.

Moreover, according to Definition 1, we have:

$$v_\Delta = \sum_{i=1}^{n} |f(x_i) - f(x_{i-1})| \leq L(|x_0 - x_1| + |x_1 - x_2| + \cdots + |x_n - x_{n-1}|) \tag{B.2}$$
$$\leq L(b-a) < \infty$$

If $V_a^x(f) < \infty$ holds, it demonstrates: $sup\{v_\Delta : \forall \Delta\} < \infty$, let $x_n$ be $x$, we have:

$$\sum_{i=1}^{n} |f(x_i) - f(x_{i-1})| \leq sup\{v_\Delta : \forall \Delta\} < \infty \tag{B.3}$$

Furthermore, if $n \to \infty$, and $\sum_{i=1}^{n}|f(x_i) - f(x_{i-1})| < \infty$ holds, obviously, we must have:

$$|f(x_i) - f(x_{i-1})| \to 0 \tag{B.4}$$

It should satisfy:

$$x_i, x_{i-1} \in B(x, \varepsilon), \forall \varepsilon > 0 \ |f(x_i) - f(x_{i-1})| \leq L|x_i - x_{i-1}| \tag{B.5}$$

Since $\forall x_i, x_{i-1} \in [a,b]$, obviously, we have:

$$f \in Lip1([a,b]) \tag{B.6}$$

***Lemma 2.3*** **(Amplitude & Variance Bounded Real Function)** $\omega_f(x_0) = \lim_{\delta \to 0} sup\{|f(x') - f(x'')| : x', x'' \in B(x_0, \delta) \subseteq [a,b]\} < \varepsilon$ is equivalent to $f \in Lip1([a,b])$.

*Proof*: If $f \in Lip1([a,b])$ holds, similarly, if $n \to \infty$, $\forall \{x_i\}_{i=1}^{n}$, we have:

$$|f(x_i) - f(x_{i-1})| \to 0 \tag{B.7}$$

Replace $x_i$ and $x_{i-1}$ by $x'$, $x''$, respectively, it satisfies:

$$\omega_f(x_0) = \lim_{\delta \to 0} \sup\{|f(x') - f(x'')|: x', x'' \in B(x_0, \delta) \subseteq [a, b]\} < M \tag{B.8}$$

If $\omega_f(x_0) = \lim_{\delta \to 0} \sup \{|f(x') - f(x'')|: x', x'' \in B(x_0, \delta) \subseteq [a, b]\} < \infty$, we assume:

$$\omega_f(x_0) = \lim_{\delta \to 0} \sup \{|f(x') - f(x'')|: x', x'' \in B(x_0, \delta) \subseteq [a, b]\} < \varepsilon \tag{B.9}$$

Obviously, given $x', x'' \in B(x_0, \delta) \subseteq [a, b]$, we have:

$$\forall \varepsilon > 0 \ |f(x') - f(x'')| < \varepsilon \tag{B.10}$$

When $\delta \to 0$, let $|x' - x''| = \frac{\varepsilon}{L}$, it also indicates:

$$|f(x') - f(x'')| < L|x' - x''| \tag{B.11}$$

***Theorem 2.1*** **(Composition of Non-Smooth Activation Functions)** Given a non-smoothed activation function $f_i$ with a single non-smooth point, denoted on $[a, b] \subseteq \mathbb{R}^1$, $f_i \in Lip1([a, b]\setminus\{x_i\})$ $i \in \mathbb{N}$. And the composition of $f_i$ and $f_j$, represented as $f_{j,i} \stackrel{\text{def}}{=} f_j(f_i(x))$, results in $\mathcal{F} \stackrel{\text{def}}{=} f_{\dots, k, \dots j.i} \in Lip1([a, b]\setminus[c, d])$, when $k \to \infty$, and $m([c, d]) \neq 0$; moreover, given $t \to \infty$, the summation as $\sum_{i=1}^{t} \mathcal{F}_t$ leads to $\sum_{i=1}^{t} \mathcal{F}_t \notin Lip1([a, b]\setminus[c', d'])$ and $m([c', d']) \neq 0$. $m(\cdot)$ represents the Lebesgue measure.

***Proof:*** At first, we discuss $k < \infty$, and we assume, $f \in Lip1([a, b]\setminus\{x_0\})$

According to Lemma 1.7 and Lemma 1.8, if $x_0 \in B(x_0, \delta)$, for we have:

$$\omega_{f_i}(x_i) = \lim_{\delta \to 0} \sup\{|f_i(x') - f_i(x'')|: x', x'' \in B(x_i, \delta) \subseteq [a, b]\} > M \tag{B.12}$$

Thus, we have $f_i$ and $f_j$ are not smooth on $B(x_i, \delta)$ and $B(x_j, \delta)$, respectively.

And for the composition, let $k = 2$,

$$\omega_{f_{j,i}}(x_i) = \lim_{\delta \to 0} \sup\{|f_j(f_i(x')) - f_j(f_i(x''))|: x', x'' \in B(x_i, \delta) \subseteq [a, b]\} \tag{B.13}$$

Let $f_i(x') = x'_j$ and $f_i(x'') = x''_j$, if $(x'_j, x''_j) \in B(x_k, \delta)$, it is easy to prove the amplitude of $f_{j,i}$ as following:

$$\omega_{f_{j,i}}(x_1) = \lim_{\delta \to 0} \sup\{|f_2(x'_j) - f_2(x''_j)|: x'_j, x''_j \in B(x_k, \delta) \subseteq B(x_j, \delta) \subseteq [a, b]\} > M \tag{B.14}$$

Naturally, we need to analyze other relations of $B(x_k, \delta)$ and $B(x_j, \delta)$; in detail, there are five situations to be discussed separately:

1). Assume, if $\forall B(x_k, \delta) \cap B(x_j, \delta) = \emptyset, i, j \in \mathbb{R}, i \neq j$, obviously, due to the same composition, we have: $B(x_k, \delta) \cap B(x_{k-1}, \delta) = \emptyset$, according to Lemma 1.3,

$$[c, d] \setminus \bigcup_{k=1}^{\infty} B(x_k, \delta) = \{\hat{x}_j\}_{j=1}^{N} \tag{B.15}$$

And,

$$m\left(\{\hat{x}_j\}_{j=1}^N\right) = 0 \tag{B.16}$$

According to Lemma 1.6, $f_j(f_i(B(x_i,\delta)) \notin Lip1(B(x_k,\delta) \cup B(x_j,\delta))$, therefore, we have:

$$f_{\cdots,k,\cdots j.i} \notin Lip1([c,d]\setminus\{\hat{x}_j\}_{j=1}^N) \tag{B.17}$$

2). Similarly, if we assume $\forall B(x_k,\delta) \cap B(x_j,\delta) \neq \emptyset$, due to the composition, we have: $B(x_k,\delta) \cap B(x_{k-1},\delta) \neq \emptyset$, according to Lemma 1.5,

$$[a,b] \supseteq \bigcup_{k=1}^{K} B(x_k,\delta) \supseteq [c,d] \tag{B.18}$$

Therefore, we can conclude:

$$f_{\cdots,k,\cdots j.i} \notin Lip1([c,d]) \tag{B.19}$$

3). Moreover, if $B(x_j,\delta) \supseteq B(x_k,\delta) \supseteq B(x_{k+1},\delta) \supseteq \cdots$, according to Lemma 1.5, we have:

$$\bigcap_{i=1}^{K} B(x_i,\delta) = \Xi \neq \emptyset \tag{B.20}$$

It means:

$$f_{\cdots,k,\cdots j.i} \in Lip1([a,b]\setminus B(x_j,\delta)) \tag{B.21}$$

Thus, similarly, we have:

$$f_{\cdots,k,\cdots j.i} \in Lip1([a,b]\setminus \Xi) \tag{B.22}$$

4). Finally, if $\cdots \supseteq B(x_{k+1},\delta) \supseteq B(x_k,\delta) \supseteq B(x_j,\delta)$,

Therefore, based on (1) and (2), we have:

$$\omega_{f_{k,\cdots,2,1}}(x_2) = \lim_{\delta \to 0} \sup\{|f_{k,\cdots,2,1}(x') - f_{k,\cdots,2,1}(x'')|: x', x'' \in [c,d]\} > M \tag{B.23}$$

It indicates:

$$f_{\cdots,k,\cdots j.i} \in Lip1([a,b]\setminus \bigcup_{i=1}^{\infty} B(x_i,\delta)) \tag{B.24}$$

5). Comprehensively, the situation includes all previously discussed (1) to (4), it is easy to conclude:

$$f_{\cdots,k,\cdots j.i} \in Lip1([a,b]\setminus [c,d] \tag{B.25}$$

Using Lemma 1.1, obviously, given $\Delta: a = x_0 < x_1 < x_2 < \cdots < x_n = b$, and $\Delta': x' < \hat{x}_1 < \hat{x}_2 < \cdots < \hat{x}_n < x''$, $v_{\Delta_1} + v_{\Delta_2} = v_\Delta$.

$$v_\Delta + v_{\Delta'} = v_{\Delta_1} + v_{\Delta_2} + v_{\Delta'} \tag{B.26}$$

$$= \sum_{i=1}^{n_1} |f(x_i) - f(x_{i-1})| + \sum_{i=n_1}^{n_2} |f(x_i) - f(x_{i-1})|$$

$$+ \sum_{i=n_2}^{n} |f(x_i) - f(x_{i-1})|$$

Since $v_{\Delta'} > M$, $v_\Delta + v_{\Delta'} > M$, it is easy to have:

$$\sum_{i=1}^{t} f_i^t \notin Lip1([a,b]) \tag{B.27}$$